\pdfoutput=1
\documentclass[10pt,twocolumn,letterpaper]{article}

\usepackage{wacv}              
\usepackage[accsupp]{axessibility}

\usepackage{graphicx}
\usepackage{amsmath}
\usepackage{amssymb}
\usepackage{booktabs}

\usepackage{multirow}

\usepackage[dvipsnames]{xcolor}
\usepackage{bbm}

\usepackage[pagebackref,breaklinks,colorlinks]{hyperref}

\usepackage[capitalize]{cleveref}
\crefname{section}{Sec.}{Secs.}
\Crefname{section}{Section}{Sections}
\Crefname{table}{Table}{Tables}
\crefname{table}{Tab.}{Tabs.}

\def\wacvPaperID{471} 
\def\confName{WACV}
\def\confYear{2024}

\begin{document}

\title{Foundation Model Assisted Weakly Supervised Semantic Segmentation}

\author{Xiaobo Yang\\
Zhejiang University\\
Hangzhou, China\\
{\tt\small hal\_42@zju.edu.cn}
\and
Xiaojin Gong\\
Zhejiang University\\
Hangzhou, China\\
{\tt\small gongxj@zju.edu.cn}
}
\maketitle

\begin{abstract}
This work aims to leverage pre-trained foundation models, such as contrastive language-image pre-training (CLIP) and segment anything model (SAM), to address weakly supervised semantic segmentation (WSSS) using image-level labels. To this end, we propose a coarse-to-fine framework based on CLIP and SAM for generating high-quality segmentation seeds. Specifically, we construct an image classification task and a seed segmentation task, which are jointly performed by CLIP with frozen weights and two sets of learnable task-specific prompts. A SAM-based seeding (SAMS) module is designed and applied to each task to produce either coarse or fine seed maps. Moreover, we design a multi-label contrastive loss supervised by image-level labels and a CAM activation loss supervised by the generated coarse seed map. These losses are used to learn the prompts, which are the only parts need to be learned in our framework. Once the prompts are learned, we input each image along with the learned segmentation-specific prompts into CLIP and the SAMS module to produce high-quality segmentation seeds. These seeds serve as pseudo labels to train an off-the-shelf segmentation network like other two-stage WSSS methods. Experiments show that our method achieves the state-of-the-art performance on PASCAL VOC 2012 and competitive results on MS COCO 2014. Code is available at \href{https://github.com/HAL-42/FMA-WSSS.git}{https://github.com/HAL-42/FMA-WSSS.git}.
\end{abstract}

\begin{figure}[htpb]
\begin{center}
\includegraphics[scale = 0.165]{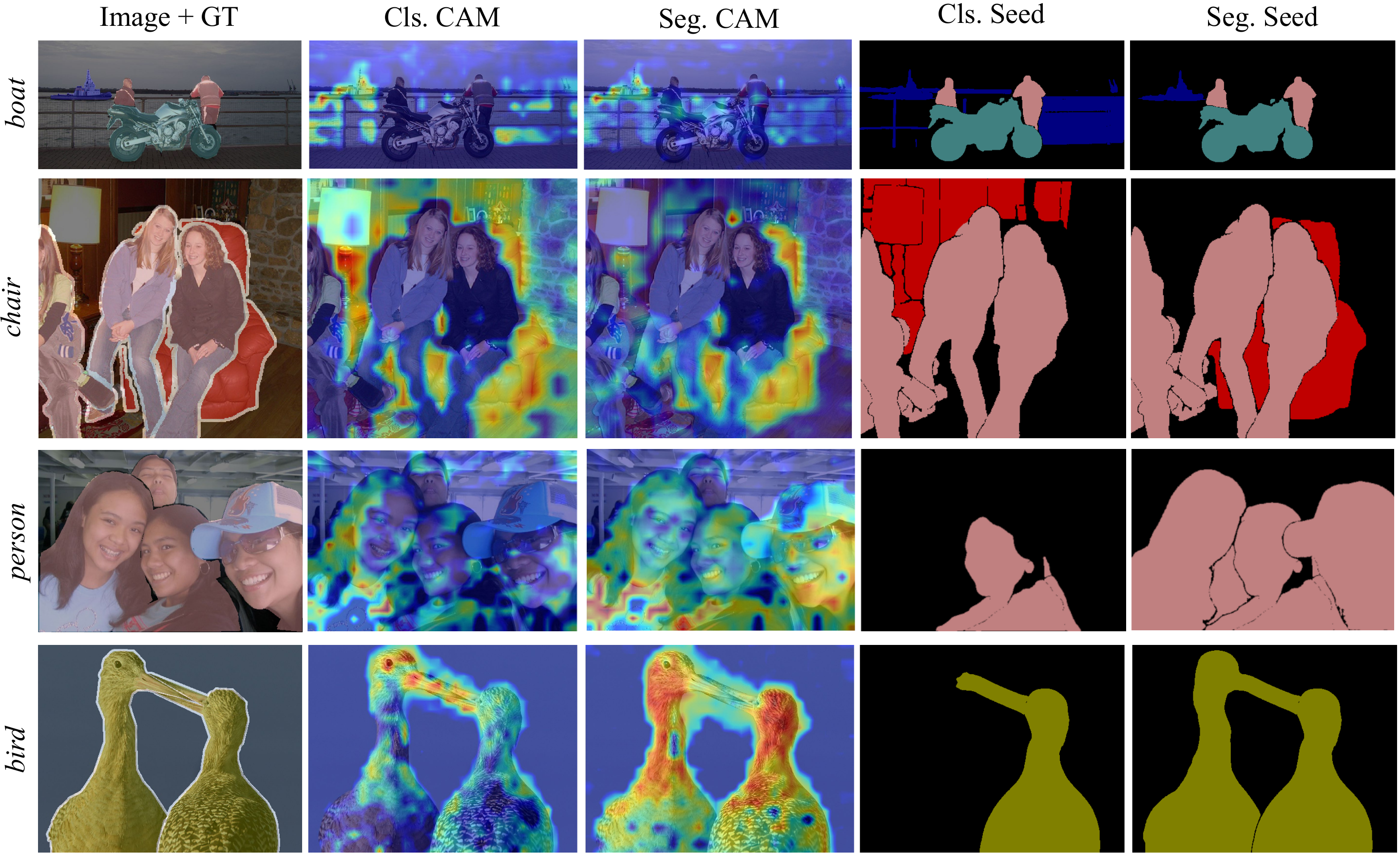}
\end{center}
\vspace{-12pt}
\caption{
Typical examples of the class activation maps (CAMs) and seed maps produced from the classification (Cls.) and segmentation (Seg.) tasks constructed in our coarse-to-fine framework. The CAMs only demonstrate the maps activated by a target class labelled on the left side, while the seed maps cover all foreground classes presented in the images. 
}
\label{fig:highlight}
\vspace{-12pt}
\end{figure}

\section{Introduction}
\label{sec:intro}
Weakly supervised semantic segmentation (WSSS) with image-level labels~\cite{Wei2017,Ahn2018AffinityNet,Xu2021,Zhou2022} has attracted a great amount of research interest due to its much lower annotation cost compared to fully supervised counterparts~\cite{Long2015fcn,chen2017deeplab,Wang2021hrnet,Xie2021}. Mainstream methods address this WSSS problem by obtaining reliable seeds and using them as pseudo labels to train an off-the-shelf segmentation network. Therefore, extensive efforts have been devoted to generating high-quality segmentation seeds. These methods often utilize class activation maps (CAMs)~\cite{Zhou2016CAM,Selvaraju2017GradCAM} derived from image classification networks as initial seeds. However, the obtained CAMs tend to focus on the most discriminative parts rather than complete objects and may falsely activate the background. To improve them, various techniques such as adversarial erasing~\cite{Wei2017,Kweon2021,Lee2021AdvCAM}, saliency guidance~\cite{Yao2020,Lee2021}, affinity learning~\cite{Ahn2018AffinityNet,Fan2020AffinityNet,Xu2021}, and contrast learning~\cite{Zhou2022,Du2022} have been exploited. Despite significant advancements achieved by these methods, the performance of WSSS still lags behind that of FSSS counterparts.

Recently, pre-trained foundation models such as the contrastive language-image pre-training (CLIP) model~\cite{Radford2021CLIP} and the segment anything model (SAM)~\cite{Kirillov2023SAM} have also been leveraged to improve the CAMs~\cite{Xie2022CLIMS,Lin2023CLIP-ES,Jiang2023SAMwsss}. While these methods~\cite{Xie2022CLIMS, Lin2023CLIP-ES, Jiang2023SAMwsss} have shown promising results, the way of using foundation models is still in its preliminary stage, leaving ample room for further exploration. For instance, in the adaptation of CLIP to WSSS, both CLIMS~\cite{Xie2022CLIMS} and CLIP-ES~\cite{Lin2023CLIP-ES} rely on manually-designed prompts, neglecting the potential benefits of using learnable prompts~\cite{Zhou2022COOP}, which have demonstrated effectiveness in other vision tasks. Furthermore, when incorporating SAM into WSSS, a challenge arises as SAM commonly supports point- or box-wise prompts and produces part/object-level masks without class labels. Previous methods either produce box-wise prompts~\cite{Sun2023SAMwsss} as an initial step or adopt simple voting schemes~\cite{Chen2023SAMwsss, Jiang2023SAMwsss} to select masks. How to effectively integrate SAM into the WSSS framework remains underexplored. 
 
This work attempts to leverage both CLIP and SAM to generate high-quality segmentation seeds for WSSS.
Firstly, we introduce prompt learning~\cite{Zhou2022COOP} to adapt CLIP more effectively. A straightforward way to use prompt learning is to replace the manually-designed prompts used in CLIP-ES~\cite{Lin2023CLIP-ES} with contextually learnable prompts trained through an image classification task.
However, due to the weak supervision of image-level labels, the \textit{coarse seeds} generated from classification using the learned prompts may still be under-perform, missing object parts or including background regions, as shown in Figure~\ref{fig:highlight}.
To improve them, we additionally construct a segmentation task and learn a set of segmentation-specific prompts relying on the coarse seeds. 
This joint framework for classification and segmentation refine the initial segmentation seeds in a progressive manner, moving from a coarse to a fine level. 
Finally, the \textit{fine seeds} produced by segmentation-specific prompts are used to train the final segmentation network.

In both classification or segmentation tasks of our coarse-to-fine framework, we keep the CLIP model frozen while only learning task-specific prompts. A multi-label contrastive loss relying on image-level labels and a CAM activation loss relying on the generated coarse seeds are designed to train the classification and segmentation tasks jointly. In addition, we employ Softmax-GradCAM~\cite{Lin2023CLIP-ES} to produce the class activation maps in each task stream. Then, instead of taking dense CRF~\cite{Krahenbuhl2011CRF} or AffinityNet~\cite{Ahn2018AffinityNet} like previous WSSS methods to refine the CAMs, we design a SAM-based seeding module for refinement, which takes advantage of class-agnositic part/object-level masks produced by SAM to generate seed maps. The CLIP-based coarse-to-fine framework, together with the SAM-based seeding module and the designed losses, empower us to generate high-quality segmentation seeds.

In summary, the contributions of our work are as follows:
\begin{itemize}
\vspace{-0.5em}
\item We propose a coarse-to-fine framework that leverages CLIP to jointly perform classification and segmentation tasks. The coarse seed map yielded from classification guides the generation of the fine seed map from segmentation. Moreover, we learn two sets of task-specific prompts, enabling the frozen CLIP model to be adapted to our two tasks more effectively than manually defined prompts. 	
\vspace{-0.5em}
\item We design a SAM-based seeding module that can be applied at the seed generation stage to produce high-quality seeds from CAMs and at the second stage to refine final segmentation results. In contrast to dense CRF~\cite{Krahenbuhl2011CRF} or AffinityNet~\cite{Ahn2018AffinityNet} that are extensively used in previous WSSS methods, our module generates better refinement results while being more efficient. 
\vspace{-0.5em}
\item Extensive experiments show that out method achieves state-of-the-art on PASCAL VOC 2012~\cite{everingham2010pascal} and very competitive results on MS COCO 2014~\cite{Lin2014COCO}.
        
\end{itemize}

\section{Related Work}
\subsection{Weakly Supervised Semantic Segmentation}
WSSS methods can be roughly grouped into one-stage and two-stage techniques. The one-stage methods~\cite{Ru2022,Araslanov2020,Zhang2020} train a segmentation network end-to-end using image-level labels. In contrast, the two-stage methods~\cite{Ahn2018AffinityNet,Kweon2021,Rong2023} first generate segmentation seeds and then use them as pseudo labels to train an off-the-shelf segmentation network, achieving better performance. A key problem in the two-stage methods is to generate high-quality seeds. To this end, various methods such as adversarial erasing~\cite{Wei2017,Kweon2021,Lee2021AdvCAM}, saliency guidance~\cite{Yao2020,Lee2021}, affinity learning~\cite{Ahn2018AffinityNet,Fan2020AffinityNet,Xu2021}, contrast learning~\cite{Zhou2022,Du2022,YangXiaoYang}, and boundary-aware~\cite{Li2022,Rong2023} techniques have been proposed to improve the seeds initially generated by CAMs~\cite{Zhou2016CAM,Selvaraju2017GradCAM}. Recently, a new trend~\cite{Lin2023CLIP-ES,Chen2023SAMwsss} is to take advantage of pre-trained large foundation models. In line with this research direction, our work has explored incorporating CLIP~\cite{Lin2023CLIP-ES} with SAM~\cite{Kirillov2023SAM} to generate high-quality seeds. 
%
 %
 %

\subsection{Contrastive Language-Image Pre-training}
Contrastive Language-Image Pre-training (CLIP)~\cite{Radford2021CLIP} is a vision-language model trained on 400 million image-text pairs collected from the Internet. It demonstrates a strong ability to generalize to unseen data and has been adapted to various downstream vision tasks. In WSSS, CLIMS~\cite{Xie2022CLIMS} proposed CLIP-based losses to supervise another network to generate high-quality CAMs. CLIP-ES~\cite{Lin2023CLIP-ES} designed Softmax-GradCAM and class-aware attention-based affinity (CAA) to generate CAMs directly from CLIP. Both methods~\cite{Xie2022CLIMS,Lin2023CLIP-ES} use manually-designed templates like ``\textit{a photo of [CLS]}'' or ``\textit{a clean origami [CLS]}'' as text prompts when leveraging textual information of CLIP. In contrast, learnable textual contexts have shown their effectiveness in achieving better transferability in other vision tasks~\cite{Zhou2022COOP,Rao2022DenseCLIP}. Inspired by them, we propose to learn two sets of prompts specific to classification and segmentation tasks in our coarse-to-fine framework, enabling the frozen CLIP model to be adapted to our tasks more effectively.

\subsection{Segment Anything Model}
The Segment Anything Model (SAM)~\cite{Kirillov2023SAM} is a recently released foundation model for image segmentation. Trained with over 1 billion masks on 11 million images, it has gained a strong generalization ability and can be easily adapted to a range of downstream vision tasks. SAM has also been applied to boost the performance of WSSS very recently. For instance, Chen \etal~\cite{Chen2023SAMwsss} improved the CAM-based pseudo labels by including SAM-generated segments based on overlap ratios. Jiang \etal~\cite{Jiang2023SAMwsss} produced pseudo segmentation labels by prompting SAM with local maximum points on the CAMs. Contrastively, we introduce a SAM-based seeding module that incorporates a filter with a preference order of \textit{whole}, \textit{part}, and \textit{subpart} to generate semantic quasi-superpixels from SAM-generated masks, based on which more complete and precise segmentation seeds are generated. A concurrent work to ours is presented in~\cite{Sun2023SAMwsss}, which employed Grounded-DINO~\cite{GroundingDINO} to generate box-wise prompts and feed them into SAM to produce segmentation seeds. In contrast to it, our approach incorporates CLIP with SAM for seed generation.

\section{Preliminary}
CLIP-ES~\cite{Lin2023CLIP-ES} is a WSSS framework that inspires our work. In this section, we briefly introduce three of its components, including text prompt selection, Softmax-GradCAM, and class-aware attention-based affinity (CAA), which are related to or will be used in our work.  

\textbf{Text prompt selection.}
CLIP-ES proposes a sharpness-based criterion to guide the choice of text prompts. The selection of prompts is made through trial and error, and ``\textit{a clean origami [CLS]}'', which has the lowest sharpness, is manually selected as the input prompt. Here, \textit{[CLS]} is a foreground class label and its synonyms, or a background label such as \textit{ground}, \textit{grass}, \textit{railroad}, etc., which are additionally defined.

\textbf{Softmax-GradCAM.} 
CLIP-ES introduces the Softmax function into the original GradCAM~\cite{Selvaraju2017GradCAM} to generate CAMs that better suppress non-target classes and backgrounds. Specifically, for each image and a text prompt of any one class label, CLIP yields an image embedding $\textbf{f}_I$ and a text embedding $\textbf{f}_T$, and a logit $Y = cos(\textbf{f}_I, \textbf{f}_T)/\tau$ is computed, where $cos(\cdot, \cdot)$ denotes cosine similarity and $\tau$ is a temperature factor. Then, for the image, a Softmax score $s^c$ for class $c\in \mathcal{P}$ is computed as follows:
\vspace{-3pt}
\begin{equation}
    s^c = \frac{\exp(Y^c)}{\Sigma_{c'\in \mathcal{P}\cup \mathcal{B}}\exp(Y^{c'})},  
\end{equation}
where $\mathcal{P} \subseteq \mathcal{F}$ denotes the set of classes presented in the image, and $\mathcal{F}$ is the set of foreground classes in a whole dataset. $\mathcal{B}$ denotes the set of background classes manually defined in CLIP-ES.

Then, given a feature map $F \in \mathbb{R}^{K \times H \times W}$ output before the last multi-head self-attention (MHSA) layer of the CLIP image encoder, the activation weight corresponding to class $c$ at $k$-th channel is calculated by
\vspace{-3pt}
\begin{equation}
    w^c_k = 
    \frac{1}{HW}
    \sum_{u=1}^H \sum_{v=1}^W
    \frac{\partial s^c}{\partial F_{k,u,v}}.
\end{equation}
The CAM of class $c$ is finally obtained by 
\vspace{-3pt}
\begin{equation}
    M^{c}_{u,v} = \text{ReLU}\left(\sum_{k=1}^K w^c_k F_{k,u,v}\right).
\end{equation}

\textbf{Class-aware attention-based affinity.} CAA is proposed to refine the above-obtained CAMs by leveraging the attention weight $W^{attn}\in \mathbb{R}^{HW \times HW}$ of the last MHSA layer. That is,
\vspace{-3pt}
\begin{equation}
    {M^c}^* = B^c \odot A^t \cdot vec(M^c),
\end{equation}
where $B^c\in \mathbb{R}^{1\times HW}$ is a box mask taken from the CAM of class $c$, $\odot$ is Hadamard product, $t$ is the iteration number, $vec(\cdot)$ is the vectorization of a matrix, and $A$ is a symmetric affinity matrix obtained from the attention weight. (Please refer to CLIP-ES~\cite{Lin2023CLIP-ES} for more details.)

\begin{figure*}[htpb]
\begin{center}
\includegraphics[scale = 0.225]{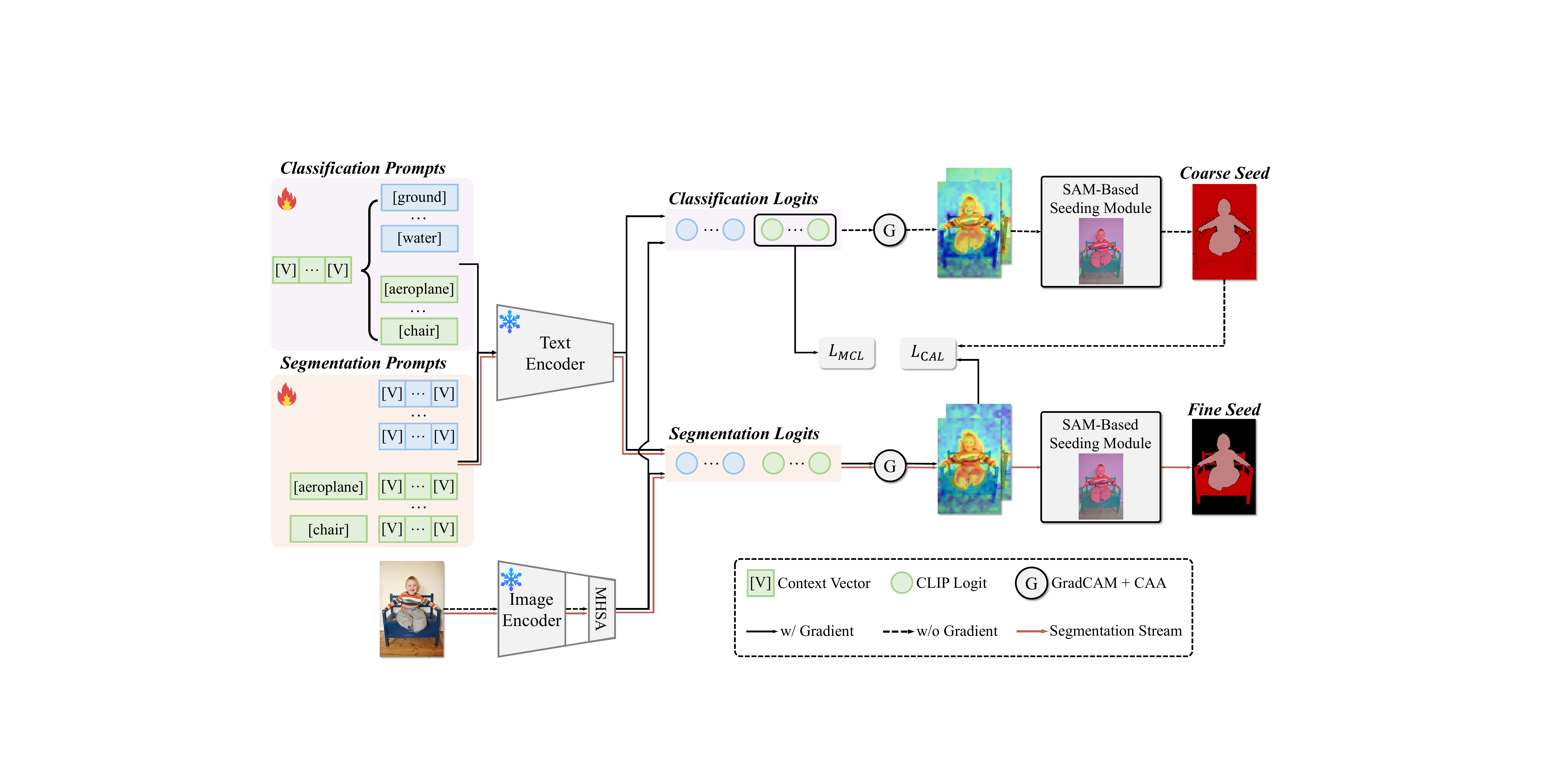}
\end{center}
\vspace{-12pt}
\caption{
An overview of the proposed coarse-to-fine framework for the generation of high-quality segmentation seeds. It employs a frozen CLIP model with two sets of learnable task-specific prompts to jointly perform image classification and seed segmentation tasks. The logits computed from each set of prompts are used by Softmax-GradCAM to calculate class activation maps (CAMs), which are further refined by class-aware attention-based affinity (CAA). The CAMs obtained from each task are then input into a SAM-based seeding module to generate either a coarse or a fine seed map. Moreover, a multi-label contrastive loss $\mathcal{L}_{MCL}$ and a CAM activation loss $\mathcal{L}_{CAL}$ are designed to supervise the learning of the prompts, which are the only parts that need to be learned in our framework. Once learned, only the segmentation stream, as shown by red lines, is utilized to generate high-quality seeds.
}
\label{fig:overview}
\vspace{-12pt}
\end{figure*} 

\section{The Proposed Method}
In this section, we present a coarse-to-fine framework based on CLIP~\cite{Radford2021CLIP} and SAM~\cite{Kirillov2023SAM} to produce high-quality segmentation seeds. For each image, a coarse seed map is generated from an image classification task that is supervised by a multi-label contrastive loss using the image-level labels. Then, a fine seed map is obtained from a seed segmentation task that is supervised by a CAM activation loss using the generated coarse seeds. We leverage a frozen CLIP model while learning task-specific prompts to perform both classification and segmentation tasks. In each task, Softmax-GradCAM and CAA~\cite{Lin2023CLIP-ES} are applied to produce class activation maps, which are fed into a SAM-based seeding module to generate either coarse or fine seeds. Figure~\ref{fig:overview} illustrates an overview of the entire framework. Once the task-specific prompts are learned, high-quality segmentation seeds will be produced using the individual segmentation stream and further used as pseudo labels to train an off-the-shelf segmentation network.

\subsection{Prompt Learning}
In contrast to CLIP-ES~\cite{Lin2023CLIP-ES} that manually selects text prompts, we employ prompt learning~\cite{Zhou2022COOP} to adapt CLIP to WSSS more effectively. Furthermore, instead of learning task-share prompts that overlook fine-grained task correlations, we learn two sets of prompts specific to image classification and seed segmentation tasks, respectively.



\textbf{Classification-specific prompts.} 
In our WSSS framework, the image-level labels are used to learn a multi-label image classification task. Therefore, we first design a set of prompts specific to this task. Considering that the image-level supervision is relatively weak, we choose a unified context strategy that shares the same context among all classes to avoid overfitting. More specifically, the unified context is defined as follows:
\begin{equation}
    t^c = [V]_1[V]_2\cdots[V]_N[CLS^c],
\end{equation}
where $[V]_n$ ($n \in \{1, ..., N\}$) is a vector with the same dimension as the word embeddings of CLIP, and $N$ is the number of context tokens. $c \in \mathcal{F}\cup \mathcal{B}$, in which $\mathcal{F}$ is the set of foreground classes given in a dataset and $\mathcal{B}$ is the set of background classes manually defined in CLIP-ES~\cite{Lin2023CLIP-ES}. This prompt learning technique enables the CLIP model to adapt effectively to our multi-label classification task.

\textbf{Segmentation-specific prompts.} 
Our framework also includes a seed segmentation task, and therefore, we design a set of segmentation-specific prompts as well. Considering that the supervision of this task is on a pixel level, which is stronger than that of the classification task, we choose a class-specific context strategy in which each class has independent context vectors. That is, for each foreground class $c \in \mathcal{F}$, we define the prompt as
\begin{equation}
t^c = [CLS^c][V]^c_1[V]^c_2\cdots[V]^c_N,
\end{equation}
and for a background class $b \in \mathcal{B}$ we define the prompt as
\begin{equation}
t^b = [V]^b_1[V]^b_2\cdots[V]^b_N,
\end{equation}
where $[V]^c_n$ and $[V]^b_n$ have the same dimension as the word embeddings as well. 

It is worth noting that, here we prepend the foreground class label to the contexts, which has been validated to be more effective compared to the way used in classification-specific prompts. Additionally, as highlighted by CLIP-ES~\cite{Lin2023CLIP-ES}, the regions identified by foreground and background prompts are implicitly mutually exclusive when employing Softmax-GradCAM. This allows us to eliminate the need for manually defined background class labels in segmentation-specific prompts and still make the background prompts to be fully learned. 
%

%
%


\subsection{SAM-based Seeding Module}
The SAM-based seeding module takes the CAMs as input to produce a seed map. By leveraging the object masks produced by the segment anything model (SAM)~\cite{Kirillov2023SAM}, we implement this module through three steps: quasi-superpixel generation, quasi-superpixel classification, and seed map generation, which are introduced below.

\textbf{SAM-based quasi-superpixel generation.}
When prompted with a regular grid of points, SAM~\cite{Kirillov2023SAM} produces $\sim$100 high-quality masks per image. However, these masks lack semantic labels, and many of them are overlapped, covering the \textit{whole}, \textit{part}, and \textit{sub-part} of objects. To leverage these masks for generating complete and precise segmentation seeds, we first screen out a set of suitable masks, which we refer to as quasi-superpixels. These quasi-superpixels are selected to preferentially cover whole objects. Additionally, similar to standard non-overlapping superpixels, these quasi-superpixels are chosen to minimize overlap as much as possible.

\begin{figure}[thpb]
\begin{center}
\includegraphics[scale = 0.105]{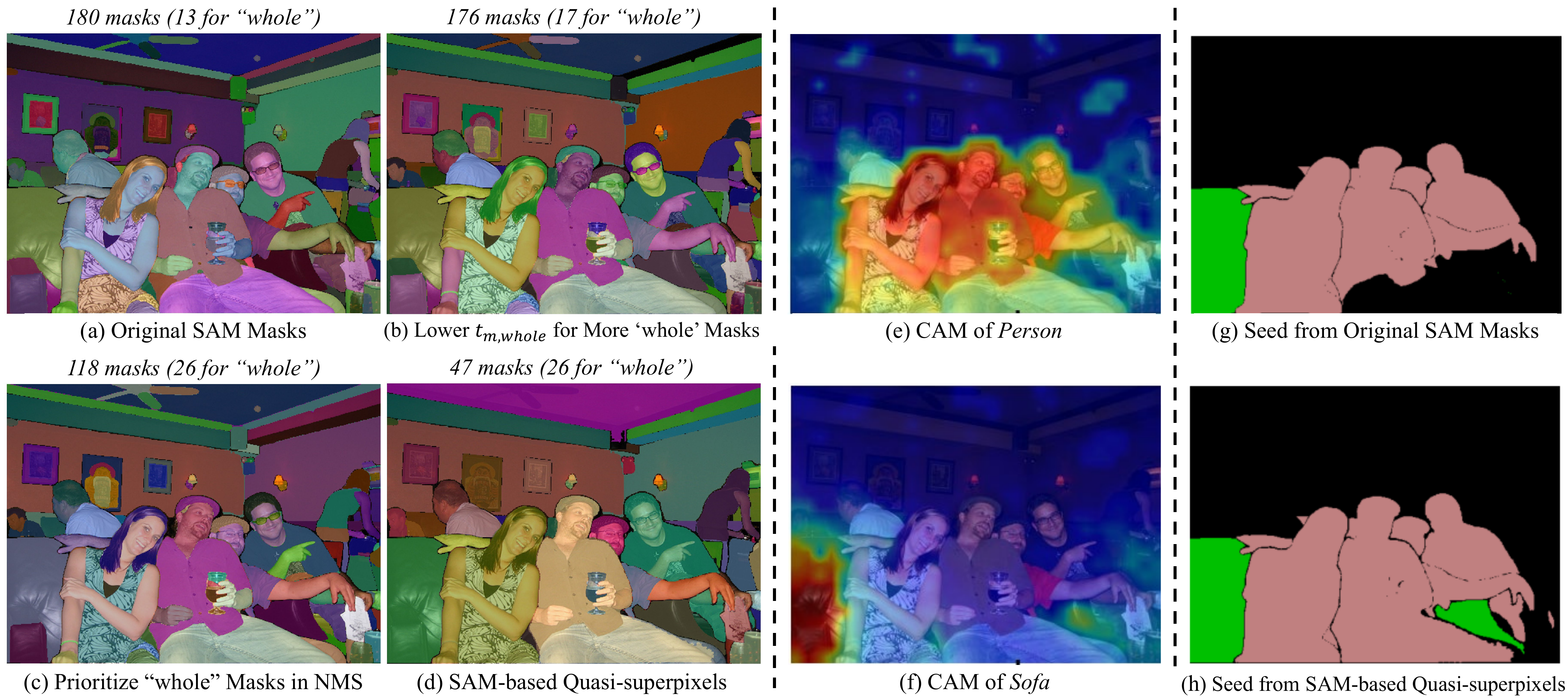}
\end{center}
\vspace{-12pt}
\caption{Illustration of the object masks generated by SAM and the resulted seed maps. 
}
\label{fig:seed}
\vspace{-12pt}
\end{figure}

To this end, we make slight modifications to the mask generation pipeline of SAM as follows: 1) We set a lower confidence threshold $t_{m,whole}$ to retrieve more masks at the \textit{whole} level. 2) During non-maximal suppression (NMS), we prioritize the selection of \textit{whole} level masks, reducing the suppression of \textit{whole} masks caused by overlaid \textit{part} or \textit{sub-part} masks. 3) We design a simple filter and apply it to the masks retained after NMS to further reduce overlap. The filter follows a preference order of \textit{whole}, \textit{part}, and \textit{sub-part}. That is, it selects \textit{whole} masks first, followed by \textit{part} masks, and then \textit{sub-part} masks. Each subsequent level mask is included only if the occupation ratio between it and the selected masks, which is the ratio between the intersection and the subsequent level mask area, is lower than a threshold $t_r$. Figure~\ref{fig:seed} demonstrates typical examples of the masks selected after each of these steps.

\textbf{SAM-based quasi-superpixel classification.}
We utilize the refined CAMs to determine the semantic class for each quasi-superpixel. The CAM for a foreground class $c$ is denoted by $M^{c*}\in \mathbb{R}^{H\times W}$, and each quasi-superpixel $Q^i$ is represented by a binary mask of size $H\times W$. The score for $Q^i$ being classified as class $c$ is computed by averaging the activation weights within the quasi-superpixel and then normalizing over all quasi-superpixels. That is,
\vspace{-3pt}
\begin{equation}
S^c(Q^i) = Norm\left(\frac{1}{|Q^i|}\sum^{H}_{u=1} \sum^{W}_{v=1} Q^i_{u,v} \cdot M^{*c}_{u,v}\right),
\end{equation} 
where $Norm(\cdot)$ is min-max normalization. The score of being classified as the background is calculated by
\begin{equation}
    S^{bg}(Q^i) = (1 - \max_{c\in \mathcal{P}}(S^{c}(Q^i)))^{\alpha},
\end{equation}
where $\alpha$ is a hyper-parameter empirically set. A larger value of $\alpha$ suppresses the background and results in more foreground superpixels.

Then, the semantic class of quasi-superpixel $Q^i$ is determined by
\begin{equation}
    l_i = \arg \max_{j\in \mathcal{P} \cup \{bg\}}(S^j(Q^i)).
\end{equation}
Note that, unlike prompt learning that considers multiple background classes, here we only consider one background class denoted as \textit{bg}.

\textbf{Seed map generation.}
Finally, we generate a seed map $D \in \mathbb{R}^{H\times W}$ by assigning the class label of each quasi-quasi-superpixel to the pixels it contains. If there are some pixels covered by two quasi-superpixels, which is rare after our filtering scheme, the class of the quasi-superpixels at the higher level is assigned. If the levels are the same, the one with the higher classification score is assigned. Additionally, any pixels not covered by any superpixels are treated as background. In contrast to the overlap ratio based voting scheme~\cite{Chen2023SAMwsss}, our \textit{whole}-preference based seed generation is able to retrieve object masks more completely and assign semantic labels more precisely.

\subsection{Training Loss}
We design a multi-label contrastive loss $\mathcal{L}_{MCL}$ and a CAM activation loss $\mathcal{L}_{CAL}$ to supervise the learning of prompts. That is, the entire training loss is defined by 
\vspace{-3pt}
\begin{equation}
 \mathcal{L} = \mathcal{L}_{MCL} + \mathcal{L}_{CAL}.
\end{equation}

\textbf{Multi-label contrastive loss.} 
Although the binary cross entropy loss is commonly used in WSSS to supervise the multi-label classification task, we observe that the values of most CLIP logits for classification fall within the saturation range of the sigmoid function, leading to inefficient training.
We hereby employ a multi-label contrastive loss to supervise the classification task, which extends the original contrastive loss of CLIP to the multi-label scenario:
\begin{equation}
\resizebox{0.85\hsize}{!}{$
    \mathcal{L}_{MCL}=
    \dfrac{1}{|\mathcal{P}|}
    \sum_{c \in \mathcal{P}}
    -\log \dfrac{\exp (Y^c_C)}
    {\exp (Y^c_C) + \sum_{c' \in \mathcal{F} - \mathcal{P}} \exp(Y^{c'}_C)},
$}
\end{equation}
in which $Y_C$ represents a logit for the classification task, $\mathcal{P}$ is the set of classes presented in the image, and $|\cdot|$ is the cardinality of a set. This contrastive loss attracts the image embedding to the text embeddings of foreground classes present in the image while pushing it apart from the text embeddings of absent classes. Note that although the background logits are not considered in this loss, the background prompts still benefit from the unified context shared with the foreground prompts.

\textbf{CAM activation loss.}
Although the pixel-wise cross entropy loss is commonly employed for segmentation tasks, it suffers from the low values of Softmax-GradCAM~\cite{Lin2023CLIP-ES} used in our task, leading to ineffective learning. Therefore, the CAM activation loss is designed, which encourages the activation maps yielded from the segmentation task to be closely aligned with the coarse seed map obtained from the classification task. Specifically, let us denote the $c$-class activation map generated from segmentation as $M_S^c$ and the coarse seed map obtained from classification as $D_C$. Then, the activation loss for the foreground regions is defined by
\vspace{-3pt}
\begin{equation}
    \mathcal{L}_{CAL}^{fg} = 
    \sum_{c \in \mathcal{P}} || D_{C}^c \odot \left(\max(D_{C}^c \odot M_S^c) - M_S^c\right)||,
\end{equation}
where $||\cdot||$ denotes the $L_1$ norm. The maximum value is detached to prevent it from being decreased. The loss for the background regions is defined by
\vspace{-3pt}
\begin{equation}
    \mathcal{L}_{CAL}^{bg} =  \sum_{c \in \mathcal{P}} || \overline{D_{C}^c} \odot M_S^{c}||,
\end{equation}
where $D_{C}^c = \mathbbm{1}(D_{C} = c)$ is the binary mask for class $c$ and $\overline{D_{C}^c}= 1 - D_{C}^c$.
Then, the entire CAM activation loss is 
\vspace{-3pt}
\begin{equation}
    \mathcal{L}_{CAL} = 
    \frac{1}
    {|\mathcal{P}|\cdot HW}(\mathcal{L}_{CAL}^{fg}+\mathcal{L}_{CAL}^{bg}).
\end{equation}

\subsection{Final Segmentation}
Once the coarse-to-fine framework is learned, we utilize our segmentation stream to generate high-quality seeds. These seeds further serve as pseudo labels to train an off-the-shelf segmentation network, such as DeeplabV3+~\cite{Chen2018deeplab} or Mask2Former~\cite{mask2former}, following other WSSS methods. In addition, after obtaining the final segmentation score map, we further apply our SAM-based seeding module but using the segmentation scores for quasi-superpixel classification, to refine the final segmentation results.

\section{Experiments}
\subsection{Datasets and Evaluation Metrics}
We evaluate our method on PASCAL VOC 2012~\cite{everingham2010pascal} and MS COCO 2014~\cite{Lin2014COCO} datasets. PASCAL VOC 2012~\cite{everingham2010pascal} provides pixel-wise annotations for 20 object classes and one background class, containing 1464, 1449 and 1456 images for training, validation and testing. Following~\cite{SBD}, an augmented set with 10,582 images is used for training. MS COCO 2014~\cite{Lin2014COCO} provides annotations for 80 object classes and one background class, containing 82,081 images for training and 40,137 images for validation. We train our network on each dataset using only image-level labels. Following common practice~\cite{Ahn2018AffinityNet,Fan2020AffinityNet,Xie2022CLIMS,Lin2023CLIP-ES}, we take the mean intersection-over-union (mIoU) criterion to evaluate performance.


\subsection{Implementation Details}
The pre-trained ViT-B/16~\cite{Dosovitskiy2021} image encoder and the transformer text encoder of CLIP~\cite{Radford2021CLIP} are adopted and their weights are frozen. The hyper-parameters are set as follows: the number of context tokens $N$ is 16, the threshold $t_r$ is set to 0.3, and $\alpha$ is 0.6. The prompts are learned with a batch size of 16 and 25 epochs on PASCAL VOC and 10 epochs on MS COCO. Other training settings, such as the optimizer, learning rate, etc., are set following CoOp~\cite{Zhou2022COOP}.

%

For the segmentation network trained in the second stage, we utilize either ResNet101-based DeepLabV3+~\cite{Chen2018deeplab} pre-trained on ImageNet-1k or Swin-L~\cite{swin}-based Mask2Former~\cite{mask2former} pre-trained on ImageNet-21k. When using DeepLabV3+, we employ the multi-scale strategy like other methods while adopting our SAM-based seeding module as the post-processing refinement. When Mask2Former is used, neither the multi-scale strategy nor any post-processing refinement is applied. Additional details and results on more backbone architectures can be found in the supplementary materials.

\begin{table}[htpb]
\begin{center}
\caption{
Comparison of the proposed method and its variants. 
``Cls.'' and ``Seg.'' represent the classficiation-specific prompts and segmentation-specific prompts learned in our coarse-to-fine framework. ``Post-processing'' denotes the step used to generate seeds from the CAMs at the seed generation stage.
}
\label{tab:prompt}
\resizebox{1.0\hsize}{!}{
\begin{tabular}{c|ccc|cc|c}
\hline
\multirow{2}{*}{Models} & \multicolumn{3}{c|}{Prompt Context} & \multicolumn{2}{c|}{Post-Processing} & \multirow{2}{*}{mIoU(\%)} \\
 & Manual & Cls. & Seg. & CRF & SAMS & \\
\hline \hline
CLIP-ES & \checkmark & & & \checkmark & & 68.7 \\
$\mathcal{M}_0$ & & \checkmark & & \checkmark & & 68.2\\
$\mathcal{M}_1$ & & \checkmark & & & \checkmark & 71.5\\
$\mathcal{M}_2$ & & \checkmark & \checkmark & \checkmark & & 71.6\\
$\mathcal{M}_3$ & & \checkmark & \checkmark & & \checkmark & 74.2\\
\hline
\end{tabular}
}
\end{center}
\vspace{-6pt}
\end{table}

\begin{table}[htpb]
\setlength{\belowcaptionskip}{-0.5cm}  
\centering
\caption{
The performance evaluation of different designs for our segmentation-specific prompts. 
``CSC''denotes the class-specific context strategy. ``Append'' refers to putting the foreground label after the context tokens, while ``Prepend'' is putting the label before the contexts. ``w/o bg.'' denotes that context tokens are not concatenated with background class labels.}
\label{tab:Prompt2}
\scalebox{1.0}{
\begin{tabular}{l|c}
\hline
Method & mIoU(\%) \\
\hline \hline
Unified Context & 72.6\\
CSC + Append + w/ bg. & 73.3 \\
CSC + Prepend + w/ bg. & 73.6\\ 
CSC + Prepend + w/o bg. & 74.2\\
\hline
\end{tabular}
}
\vspace{-10pt}
\end{table}

\subsection{Ablation Studies}
To validate the effectiveness of our designs, we conduct a series of experiments on the PASCAL VOC 2012 \textit{trainaug} set, unless stated otherwise.
%

\textbf{Effectiveness of the coarse-to-fine framework.} To investigate the effectiveness of our coarse-to-fine framework that includes both classification and segmentation tasks, we compare it to model variants that only involve a classification task. The results are presented in Table~\ref{tab:prompt}. By comparing $\mathcal{M}_0$ to $\mathcal{M}_2$ or $\mathcal{M}_1$ to $\mathcal{M}_3$, we observe a significant performance boost when incorporating the additional segmentation task.

\textbf{Effectiveness of the prompt learning.} In our framework, we learn two sets of task-specific prompts to adapt CLIP to our two tasks. Table~\ref{tab:prompt} demonstrates that, when only the classification task is considered, the learning of prompts does not improve or even slightly degrades the performance, as seen in the comparison between CLIP-ES~\cite{Lin2023CLIP-ES} and $\mathcal{M}_0$. However, the learning of both classification-specific and segmentation-specific prompts achieves a remarkable boost in performance, as shown by $\mathcal{M}_2$.

 \begin{table}[htpb]
\centering
\caption{
The effects of refining the final segmentation results using SAMS or CRF on PASCAL VOC 2012 \textit{val} set.
}
\label{tab:SAMS Seg}
\scalebox{0.9}{
\begin{tabular}{ll|c}
\hline
Backbone & Post-Processing & mIoU(\%) \\
\hline \hline
\multirow{3}{*}{DeepLab V3+} & - & 75.9 \\
 & CRF & 75.5 \\
 & SAMS & 77.3 \\
\hline
\multirow{3}{*}{Mask2Former} & - & 82.6 \\
 & CRF & 81.4 \\
 & SAMS & 82.2 \\
\hline
\end{tabular}
}
\vspace{-6pt}
\end{table}
\begin{table}[htpb]
\centering
\caption{
The performance evaluation of different ways to generate seeds from the CAMs derived by Softmax-GradCAM. 
``Baseline'' refers to generating seeds directly from the obtained CAMs. ``CAA'' generate seeds from the CAMs refined by CAA and ``Voting'' is the method described in~\cite{Chen2023SAMwsss}. For our SAMS module, ``More \textit{whole}'' denotes setting a lower threshold for retrieving more \textit{whole} masks. ``NMS Priority'' is prioritizing \textit{whole} masks during NMS and ``Filter'' denotes the additional filtering step.
}
\label{tab:SAMS}
\scalebox{0.85}{
\begin{tabular}{ll|c}
\hline
\multicolumn{2}{l|}{Method} & mIoU(\%) \\
\hline \hline
\multicolumn{2}{l|}{Baseline} & 60.7 \\
\hline
\multicolumn{2}{l|}{+CAA} & 69.1 \\
\hline
\multicolumn{2}{l|}{+Voting~\cite{Chen2023SAMwsss}} & 69.5 \\
\multicolumn{2}{l|}{+CAA+Voting~\cite{Chen2023SAMwsss}} & 72.3 \\
\hline
\multirow{4}{*}{+SAMS} & - & 71.0 \\
 & + More \textit{whole} & 71.2 \\
 & + NMS Priority & 71.5 \\
 & + More \textit{whole} + NMS Priority & 72.0 \\ 
 & + More \textit{whole} + NMS Priority + Filter & 73.5\\
\hline
\multicolumn{2}{l|}{+CAA+SAMS} & 74.2 \\
\hline
\end{tabular}
}
\vspace{-12pt}
\end{table}

We further investigate the effectiveness of our segmentation-specific prompt design. Specifically, we study the impact of the class-specific context (CSC) strategy versus the unified context strategy, the use of the prepending scheme (\ie put a foreground label before the context tokens) versus the appending scheme (\ie put the foreground label after the contexts), as well as the inclusion or exclusion of manually defined background labels in the background prompts. The comparison results are presented in Table~\ref{tab:Prompt2}, validating that the prompt design we adopt in our framework achieves the best performance.

\textbf{Effectiveness of the SAM-based seeding module.}
The SAM-based seeding module (SAMS) can be applied as a post-processing step either at the seed generation stage to generate seeds from the CAMs or at the final segmentation stage to refine the segmentation results. In contrast, many previous methods~\cite{Lin2023CLIP-ES} utilize dense CRF~\cite{Krahenbuhl2011CRF} for post-processing. To evaluate the effectiveness of our SAMS, we compare it to CRF at both stages and present the results in Table~\ref{tab:prompt} and Table~\ref{tab:SAMS Seg}, respectively. Table~\ref{tab:prompt} demonstrates that SAMS significantly outperforms CRF when applied to the seed generation stage. Additionally, Table~\ref{tab:SAMS Seg} shows that our SAMS brings a considerable improvement, while CRF degrades the performance when applied to the final segmentation stage using DeepLab V3+.

\begin{table}[htpb]
\centering
\caption{
The performance evaluation of using different classification and segmentation losses. 
$\mathcal{L}_{BCE}$ stands for the binary cross entropy loss for classification. $\mathcal{L}_{CE}$ represents the pixel-wise cross entropy loss whose probabilities are obtained by inputting CAM activation values into a sigmoid function.
}
\label{tab:Loss Design}
\scalebox{0.9}{
\begin{tabular}{cc|cc|c}
\hline
$\mathcal{L}_{BCE}$ & $\mathcal{L}_{MCL}$ & $\mathcal{L}_{CE}$ & $\mathcal{L}_{CAL}$& mIoU(\%) \\
\hline \hline
\checkmark & & & & 65.9 \\
 & \checkmark & & &  71.5 \\
 & \checkmark & \checkmark & & 52.1 \\
 & \checkmark & & \checkmark &  74.2 \\
\hline
\end{tabular}
}
\vspace{-6pt}
\end{table}

\begin{table}[htbp]
\centering
\caption{
Quality comparison of the CAMs and segmentation seeds generated by different methods on the PASCAL VOC 2012 \textit{train} set. ``Post.'' refers to the post-processing techniques used to generate seeds from the obtained CAMs, including dense CRF (CRF), training affinity networks (RW), using class-aware attention-based affinity (CAA) and leveraging SAM. 
%
}
\label{tab:VOC Seed SOTA}
\scalebox{0.825}{
\begin{tabular}{llcc}
\hline
Methods & Post. & CAM & Seed \\ 
\hline
\hline
AdvCAM{$_{\text{\color{gray}{CVPR21}}}$}~\cite{Lee2021AdvCAM} & RW+CRF & 55.6 & 69.9 \\
MCTformer{$_{\text{\color{gray}{CVPR22}}}$}~\cite{Xu2022} & RW+CRF & 61.7 & 69.1 \\
CLIMS{$_{\text{\color{gray}{CVPR22}}}$}~\cite{Xie2022CLIMS} & RW+CRF & 56.6 & 70.5 \\
ViT-PCM{$_{\text{\color{gray}{ECCV22}}}$}~\cite{ViT-PCM} & CRF & \textbf{67.7} & 71.4 \\
AdvCAM+W-OoD{$_{\text{\color{gray}{CVPR22}}}$}~\cite{Lee2022} & RW+CRF & 59.1 & 72.1 \\
CLIP-ES{$_{\text{\color{gray}{CVPR23}}}$}~\cite{Lin2023CLIP-ES} & CAA+CRF & 58.6 & 75.0 \\

Jiang \etal{$_{\text{\color{gray}{arXiv23}}}$}~\cite{Jiang2023SAMwsss} & SAM & 47.1 & 61.9 \\



WeakTr{$_{\text{\color{gray}{arXiv23}}}$}~\cite{WeakTr} & CRF & 66.2 & 68.7 \\
\hline
Ours & CAA+SAM & 62.6 & \textbf{80.4} \\
\hline
\end{tabular}
}
\vspace{-1em} 
\end{table}

Within the SAMS module, three modifications are made to the mask generation pipeline of SAM. We here investigate the effectiveness of each modification. Moreover, segmentation seeds can also be generated directly from the CAMs derived from Softmax-GradCAM, or from the CAMs refined by CAA, or from a SAM-based voting scheme simply relied on overlap ratios as described in~\cite{Chen2023SAMwsss}. Table~\ref{tab:SAMS} presents the results obtained using all these different methods and their combinations. We observe that the best performance is achieved by using our proposed SAMS module to generate seeds from the CAMs refined by CAA.

\textbf{Effectiveness of the training loss.} We propose a multi-label contrastive loss and a CAM activation loss to learn the task-specific prompts in our framework. 
Here, we compare them with the binary cross entropy loss commonly used in the classification task and the pixel-wise cross entropy loss commonly used in the segmentation task. 
The results are presented in Table~\ref{tab:Loss Design}, demonstrating that both the multi-label contrastive and CAM activation loss outperform their counterparts by a significant margin. 

%


\begin{figure*}[htpb]
\begin{center}
\includegraphics[scale = 0.255]{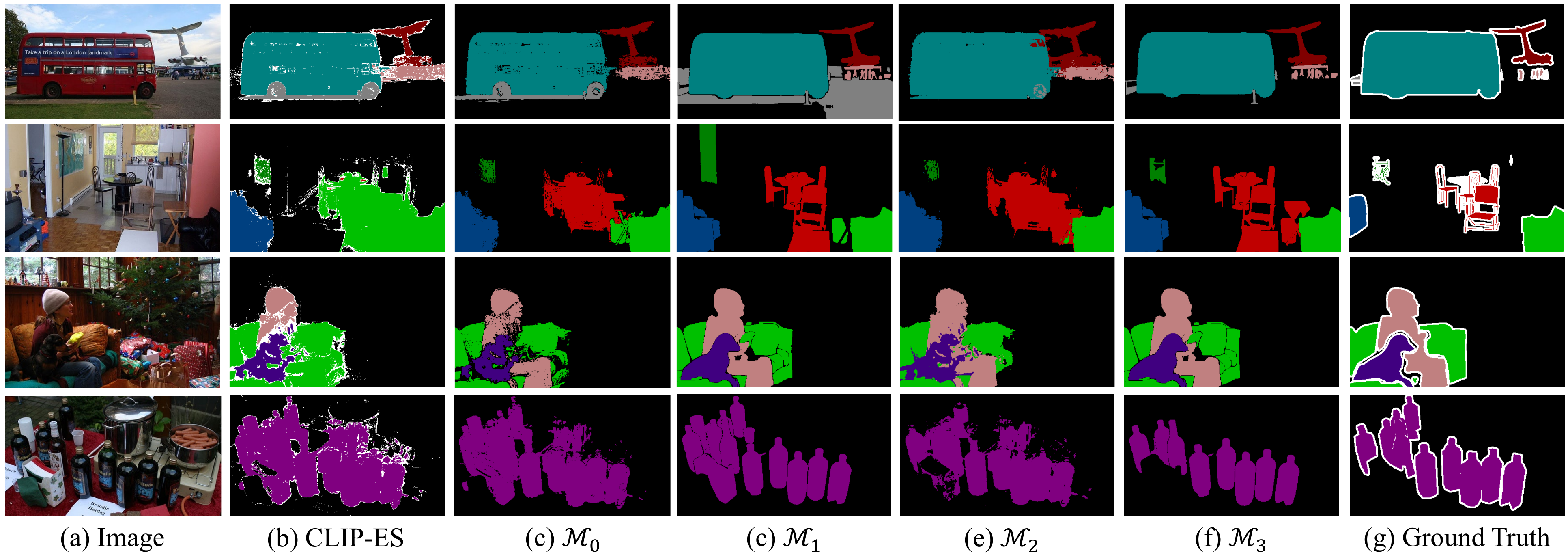}
\end{center}
\vspace{-12pt}
\caption{
Illustration of segmentation seeds generated by different models. 
}
\label{fig:voc_seeds}
\vspace{-12pt}
\end{figure*} 

\textbf{Visualization of segmentation seeds.} 
To make a qualitative comparison, Figure~\ref{fig:voc_seeds} illustrates some segmentation seeds generated by the models defined in Table~\ref{tab:prompt}. Compared to CLIP-ES~\cite{Lin2023CLIP-ES} and other model variants, our proposed full model $\mathcal{M}_3$ generates seeds with more complete regions and precise boundaries for foreground objects. This demonstrates the superior performance of our approach in seed generation for WSSS.


%
%
%
%
%
%
%
%

\begin{table}[htbp]
\centering
\caption{
Performance comparison on PASCAL VOC 2012 \textit{val} and \textit{test} sets. 
The type of supervision used for training is denoted in the "Sup." column, including full supervision (F) and image-level labels (I). The use of CLIP (C), Grounding DINO (D)~\cite{GroundingDINO}, and SAM (S) are also denoted. The \dag and \ddag indicate backbone pre-trained on COCO ground-truth~\cite{Lin2014COCO} or ImageNet-21k~\cite{ImageNet-21k}. Best WSSS results are marked in bold.
}
\label{tab:VOC SOTA}
\scalebox{0.8}{
\begin{tabular}{llccc}
\hline
Methods & Backbone & Sup. & Val & Test \\ 
\hline
\hline
DeepLab V3+{$_{\text{\color{gray}{CVPR18}}}$}~\cite{Chen2018deeplab} & R101 & F & 79.5 & 79.4 \\
Mask2Former{$_{\text{\color{gray}{CVPR22}}}$}~\cite{mask2former} & Swin-L\ddag & F & 86.0 & 86.1 \\
\hline
CIAN{$_{\text{\color{gray}{AAAI20}}}$}~\cite{Fan2020AffinityNet} & R101 & I  & 64.3 & 65.3 \\
AdvCAM{$_{\text{\color{gray}{CVPR21}}}$}~\cite{Lee2021AdvCAM} & R101 &  I  & 67.5 & 67.1 \\
Kweon \etal{$_{\text{\color{gray}{ICCV21}}}$}~\cite{Kweon2021} & WR38 &  I  & 68.4 & 68.2 \\
SIPE{$_{\text{\color{gray}{CVPR22}}}$}~\cite{Chen2022SPIE} & R101\dag &  I  & 68.8 &69.7 \\
ViT-PCM{$_{\text{\color{gray}{ECCV22}}}$}~\cite{ViT-PCM} & R101 &  I  & 70.3 & 70.9 \\
CLIMS{$_{\text{\color{gray}{CVPR22}}}$}~\cite{Xie2022CLIMS} & R50\dag &  I + C & 70.4 & 70.0 \\
AdvCAM+W-OoD{$_{\text{\color{gray}{CVPR22}}}$}~\cite{Lee2022} & WR38 &  I  & 70.7 & 70.1 \\
MCTformer{$_{\text{\color{gray}{CVPR22}}}$}~\cite{Xu2022} & WR38 &  I  & 71.9 &71.6 \\
ToCo{$_{\text{\color{gray}{CVPR23}}}$}~\cite{ToCo} & ViT-B\ddag &  I  & 71.1 & 72.2 \\
Xu \etal{$_{\text{\color{gray}{CVPR23}}}$}~\cite{Xu2023} & WR38 &  I + C & 72.2 &72.2 \\
OCR+MCTformer{$_{\text{\color{gray}{CVPR23}}}$}~\cite{Cheng2023} & WR38 &  I  & 72.7 & 72.0 \\
BECO{$_{\text{\color{gray}{CVPR23}}}$}~\cite{Rong2023} & MiT-B2 &  I  & 73.7 &73.5 \\
CLIP-ES{$_{\text{\color{gray}{CVPR23}}}$}~\cite{Lin2023CLIP-ES} & R101\dag & I + C & 73.8 & 73.9 \\
Jiang \etal{$_{\text{\color{gray}{arXiv23}}}$}~\cite{Jiang2023SAMwsss} & R101\dag & I + S &71.1 & 72.2 \\
Sun \etal{$_{\text{\color{gray}{arXiv23}}}$}~\cite{Sun2023SAMwsss} & R101\dag & I + D + S & 77.2 & 77.1 \\
WeakTr{$_{\text{\color{gray}{arXiv23}}}$}~\cite{WeakTr} & DeiT-S &  I & 74.0 & 74.1 \\
WeakTr{$_{\text{\color{gray}{arXiv23}}}$}~\cite{WeakTr} & ViT-S\ddag &  I  & 78.4 & 79.0 \\
\hline
Ours & R101 & I + C + S & 77.3 & 76.7 \\
Ours & Swin-L\ddag & I + C + S & \textbf{82.6} & \textbf{81.6} \\
\hline
\end{tabular}
}
\vspace{-1em} 
\end{table}

\vspace{-0.5em} 
\subsection{Comparison to State-of-the-Art}
\textbf{Quality of generated CAMs and segmentation seeds.} Table~\ref{tab:VOC Seed SOTA} compares the quality of the CAMs and segmentation seeds generated by our model and other methods. The results show that our model produces segmentation seeds better than all previous methods. The high-quality seeds help to train a better segmentation network for WSSS.

\begin{table}[htbp]
\centering
\caption{
Performance comparison on MS COCO 2014 \textit{val} set. 
}
\label{tab:COCO SOTA}
\scalebox{0.8}{
\begin{tabular}{llcc}
\hline
Methods & Backbone & Sup. & Val\\ 
\hline
\hline
DeepLab V3+{$_{\text{\color{gray}{CVPR18}}}$}~\cite{Chen2018deeplab} & R101 & F & 60.4\\
Mask2Former{$_{\text{\color{gray}{CVPR22}}}$}~\cite{mask2former} & Swin-L\ddag & F & 66.7\\
\hline
\hline
Kweon \etal{$_{\text{\color{gray}{ICCV21}}}$}~\cite{Kweon2021} & WR38 &  I & 36.4 \\
AdvCAM{$_{\text{\color{gray}{CVPR21}}}$}~\cite{Lee2021AdvCAM} & R101 &  I  & 44.4 \\
SIPE{$_{\text{\color{gray}{CVPR22}}}$}~\cite{Chen2022SPIE} & R101 &  I  & 40.6 \\
MCTformer{$_{\text{\color{gray}{CVPR22}}}$}~\cite{Xu2022} & WR38 &  I  & 42.0 \\
ViT-PCM{$_{\text{\color{gray}{ECCV22}}}$}~\cite{ViT-PCM} & R101 &  I  & 45.0 \\
ToCo{$_{\text{\color{gray}{CVPR23}}}$}~\cite{ToCo} & ViT-B\ddag &  I  & 42.3 \\
OCR+MCTformer{$_{\text{\color{gray}{CVPR23}}}$}~\cite{Cheng2023} & WR38 &  I  & 42.5 \\
BECO{$_{\text{\color{gray}{CVPR23}}}$}~\cite{Rong2023} & R101 &  I  & 45.1 \\
CLIP-ES{$_{\text{\color{gray}{CVPR23}}}$}~\cite{Lin2023CLIP-ES} & R101 & I + C & 45.4 \\
Xu \etal{$_{\text{\color{gray}{CVPR23}}}$}~\cite{Xu2023} & WR38 & I + C & 45.9 \\
WeakTr{$_{\text{\color{gray}{arXiv23}}}$}~\cite{WeakTr} & DeiT-S & I & 46.9 \\
WeakTr{$_{\text{\color{gray}{arXiv23}}}$}~\cite{WeakTr} & ViT-S\ddag & I  & 50.3 \\
Sun \etal{$_{\text{\color{gray}{arXiv23}}}$}~\cite{Sun2023SAMwsss} & R101 & I + D + S & \textbf{55.6} \\
\hline
Ours & R101 & I + C + S  & 48.6 \\
Ours & Swin-L\ddag & I + C + S  & 55.4 \\
\hline
\end{tabular}
}
\vspace{-1em} 
\end{table}

\textbf{Quality of final segmentation results.} We further compare the final segmentation results obtained by our method with other state-of-the-art methods on both PASCAL VOC and MS COCO datasets. The results are presented in Table~\ref{tab:VOC SOTA} and Table~\ref{tab:COCO SOTA}. When using the ResNet101 backbone on PASCAL VOC, the proposed method performs comparably to Sun \etal~\cite{Sun2023SAMwsss}, which uses a COCO-pretrained backbone. Moreover, both of them outperform other ResNet101-based methods by a significant margin. When the transformer-based backbone is utilized, our methods outperform all other methods. On MS COCO, our method also outperforms all the other methods except Sun \etal~\cite{Sun2023SAMwsss} that leverages Grounding DINO~\cite{GroundingDINO} to generate bounding boxes to prompt SAM. In contrast to CLIP~\cite{Radford2021CLIP} that is trained with text-image pairs, Grounding DINO~\cite{GroundingDINO} uses fine-grained \textit{phrase-object} pairs for training, leading to richer prior knowledge and better performance.

\vspace{-0.4em} 
\section{Conclusion}
\vspace{-0.25em}
In this work, we have presented a coarse-to-fine framework based on both CLIP and SAM to generate segmentation seeds for WSSS. The proposed task-specific prompt learning, together with the SAM-based seeding module and the designed training losses, enable our approach to generate seeds at high quality. In the proposed framework, the only parts that need to be learned are the two sets of task-specific prompts, and therefore the learning is efficient. Experimental results on PASCAL VOC and MS COCO datasets have validated the effectiveness of our method.

\clearpage


 
{\small
\bibliographystyle{ieee_fullname}
\bibliography{wacv2024_fm+wsss_v3}
}

\clearpage

\part*{Appendix}

\appendix   
\setcounter{table}{0}   
\setcounter{figure}{0}
\renewcommand{\thefigure}{S\arabic{figure}}
\renewcommand{\thetable}{S\arabic{table}}
\renewcommand{\thesection}{\Alph{section}}

\section{Illustration of SAM-based Quasi-superpixel Classification and Seed Generation}

\begin{figure}[htpb]
\begin{center}
\includegraphics[scale = 0.2]{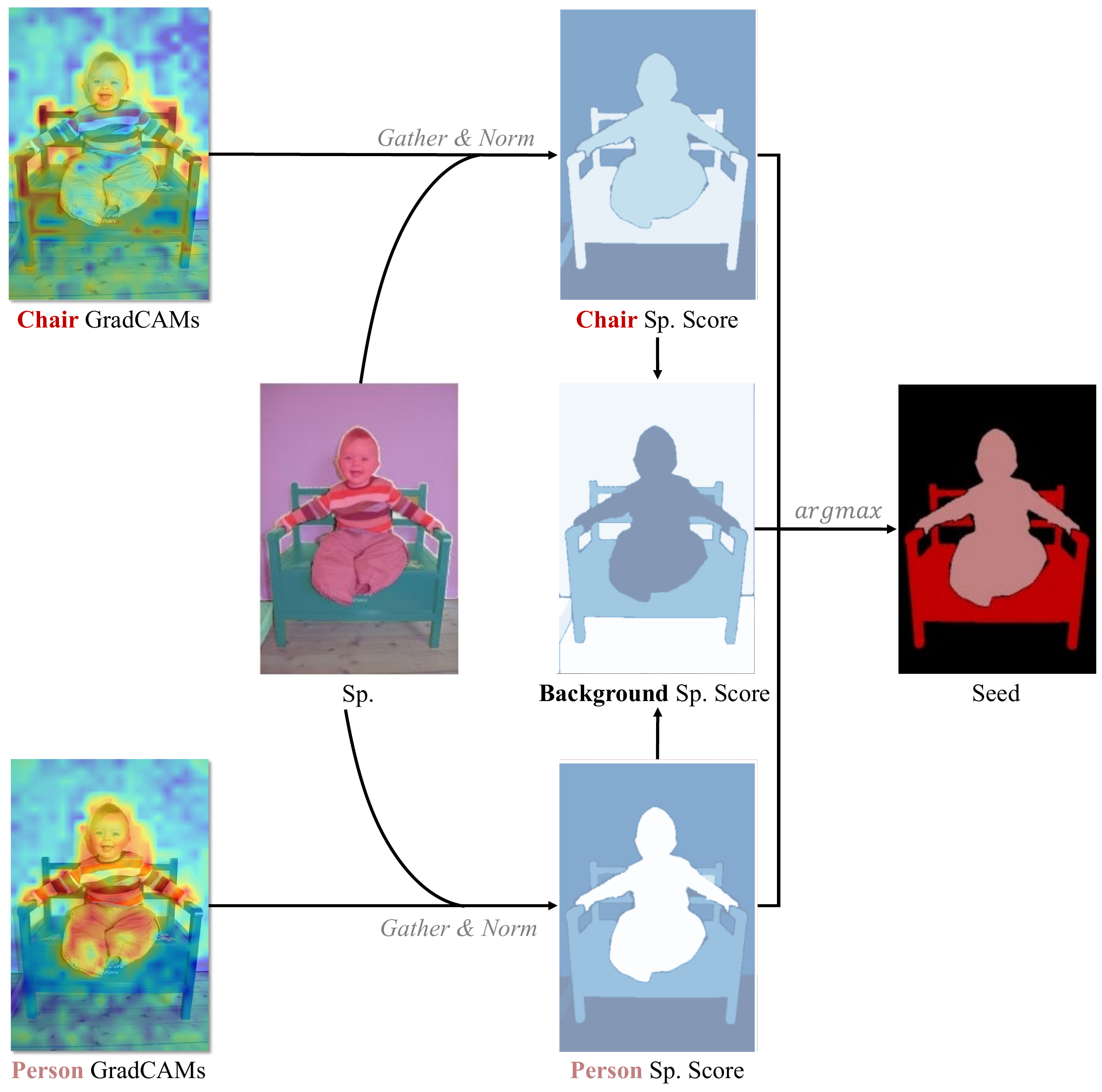}
\end{center}
\caption{
Procedures for SAM-based quasi-superpixel classification and seed map generation, where ``Sp." is the abbreviation of ``quasi-superpixel". On quasi-superpixel score maps, brighter colors indicate higher scores.
}
\label{fig:SAMS}
\end{figure} 

We illustrate the computation process of SAM-based quasi-superpixel classification and seed map generation in figure~\ref{fig:SAMS}, which depicts gathering quasi-superpixel foreground scores from GradCAMs, computing background score and determining the semantic class of each quasi-superpixel and pixel.
Please refer to Section~4.2 of the paper for more algorithm details.

\section{Framework Efficiency}

\begin{table*}[!ht]
\centering
\caption{
Time and learnable parameters size of different methods for seed generation on PASCAL VOC 2012 \textit{trainaug} set with 10,582 images.
We report the size of learnable parameters in the ``Param. Size" column.
Post-processing refinement methods include applying dense CRF (CRF), training affinity networks (RW), using class-aware attention-based affinity (CAA), and utilizing our SAM-based seeding module (SAMS).
The time unit is hour and the parameter size unit is MB. 
For MCTformer, the inference and CRF processes are combined.
}
\label{tab:Time}
\begin{tabular}{l|cc|cccc|c|c}
\hline

\multirow{2}{*}{Method} & \multicolumn{2}{c|}{CAM Generation} & \multicolumn{4}{c|}{Post-Processing} & \multirow{2}{*}{Total Time} & \multirow{2}{*}{Param. Size} \\
 & Train & Inference & CAA & SAMS & CRF & RW & & \\
\hline
\hline
CLIMS~\cite{Xie2022CLIMS} & 2.1 & 0.3 & - & - & 0.2 & 6.5 & 9.1 & 183M\\
MCTformer~\cite{Xu2022} & 0.5 & 2.5 & - & - & - & 3.0 & 6.0 & 121M\\
CLIP-ES~\cite{Lin2023CLIP-ES} & - & 0.4 & 0.01 & - & 0.2 & - & 0.6 & -\\
Ours & 1.9 & 0.4 & 0.01 & 0.01 & - & - & 2.3 & 0.8M\\
\hline
\end{tabular}
\end{table*}

In Table~\ref{tab:Time}, we compare our time costs and parameter size with some related works. 
Our method has the fewest learnable parameters except for CLIP-ES~\cite{Lin2023CLIP-ES}, which is our training-free baseline. 
Similarly, our time spent on CAM generation is minimal except for CLIP-ES. 
Furthermore, our post-processing refinement cost is negligible compared to all other methods, which is a useful feature for online refinement during training.
Note that we did not include the time required for SAM-based quasi-superpixel generation, as superpixels only need to be generated once for each training set and can be reused for all subsequent experiments.
Compared to CLIP-ES, we only consume a little additional training time and parameter size. However, in return, we achieve a considerable performance boost and eliminate the manual selection of prompt context.

\section{More Final Segmentation Results}

In this section, we trained additional combinations of segmentation networks and backbones using the fine seeds generated by our framework. We present the results on PASCAL VOC~\cite{everingham2010pascal} and MS COCO~\cite{Lin2014COCO} in Tables~\ref{tab:VOC MORE} and~\ref{tab:COCO MORE}, which also include the results from Tables~7 and~8 of the paper.

For PASCAL VOC, there is a slight performance difference between DeepLab V2~\cite{chen2017deeplab} and DeepLab V3+~\cite{Chen2018deeplab}. Switching from V3+ to V2 does not affect the conclusions drawn in Section 5.4 of the paper. 
Mask2Former~\cite{mask2former} using a larger pre-train dataset and a heavier backbone leads to a few improvements on PASCAL VOC.
Additionally, the experiments conducted on MS COCO demonstrate that Swin-B~\cite{swin} and Swin-L perform similarly, while pretraining on ImageNet-21K~\cite{ImageNet-21k} significantly improves the performance.

\section{Text-to-Semantic-Mask Usage}

SAM~\cite{Kirillov2023SAM} attempts to generate masks using CLIP text features as prompts (denoted as text-to-mask). However, its performance is not satisfactory. One solution to make SAM accept text prompts is combining Grounding-DINO~\cite{GroundingDINO}, which obtains object bounding boxes based on text inputs and then uses the object boxes to prompt SAM and generate instance masks.
We noticed that the seed generation networks in WSSS, such as CLIP-ES~\cite{Lin2023CLIP-ES} and ViT-PCM~\cite{ViT-PCM}, can be seen as text-to-semantic-mask methods for the specific data domain.
Such methods require training with data that have image-level labels. During inference, the seed generation network is prompted by text (class label) and obtains semantic segmentation masks.
We report the performance of our seed generation framework for text-to-semantic-mask in Table~\ref{tab:text-to-semantic-mask}. We also compare adopting CLIP-ES~\cite{Lin2023CLIP-ES} for text-to-semantic-mask. As can be seen, our method achieves much higher performance at the cost of additional training and inference of SAM.

Compared to combining Grounding-DINO~\cite{GroundingDINO} with SAM, our method that combines prompt-learnable CLIP has some limitations, such as not supporting free-form text input, cannot perform instance segmentation, and requiring fine-tuning on specific domains. However, on the other hand, CLIP~\cite{Radford2021CLIP} supports more semantic classes compared to Grounding-DINO. Furthermore, when fine-tuning is required for new classes or specific data distributions, our framework only requires image-level annotations rather than object box annotations. 
We hope to explore this further in future work.

\begin{table}[htpb]
\centering
\caption{
Performance comparison of our method with different final segmentation networks and backbones on PASCAL VOC 2012 \textit{val} sets. We denote segmentation network type at ``Seg." column.
The \ddag~indicates backbone pretrained on ImageNet-21k~\cite{ImageNet-21k}.
}
\label{tab:VOC MORE}
\scalebox{1.0}{
\begin{tabular}{ll|c}
\hline
Seg. & Backbone & mIoU \\ 
\hline
\hline
DeepLab V2 & R101 & 76.7 \\
DeepLab V3+ & R101 & 77.3 \\
Mask2Former & Swin-B & 80.3 \\
Mask2Former & Swin-B\ddag & 81.4 \\
Mask2Former & Swin-L\ddag & 82.6 \\
\hline
\end{tabular}
}
\end{table}

\begin{table}[htpb]
\centering
\caption{
Performance comparison of our method with different final segmentation networks and backbones on MS COCO 2014 \textit{val} set. We denote segmentation network type at ``Seg." column.
The \ddag~indicates backbone pretrained on ImageNet-21k~\cite{ImageNet-21k}.
}
\label{tab:COCO MORE}
\scalebox{1.0}{
\begin{tabular}{ll|c}
\hline
Seg. & Backbone & mIoU \\ 
\hline
\hline
DeepLab V3+ & R101 & 48.6 \\
Mask2Former & Swin-B & 51.8 \\
Mask2Former & Swin-B\ddag & 55.1 \\
Mask2Former & Swin-L\ddag & 55.4 \\
\hline
\end{tabular}
}
\end{table}

\vspace{-0.15em}
\section{More Ablation Studies on Training Loss}

\begin{table}[htbp]
\centering
\caption{
Performance comparison of text-to-semantic-mask usage on PASCAL VOC 2012 \textit{val} sets. 
}
\label{tab:text-to-semantic-mask}
\scalebox{1.0}{
\begin{tabular}{l|c}
\hline
Method & mIoU \\ 
\hline
\hline
CLIP-ES~\cite{Lin2023CLIP-ES} & 73.8 \\
Ours & 80.6 \\
\hline
\end{tabular}
}
\end{table}
The values of most CLIP logits are situated in the saturated range of the sigmoid, leading to inefficient training by binary cross entropy loss.
Moreover, we use positive class Softmax-GradCAMs with sigmoid activation to calculate probabilities for pixel-wise cross entropy loss. 
Note that we only obtain the Softmax-GradCAMs of positive classes, so probabilities can only be derived from sigmoid rather than softmax.
However, due to CLIP parameters being frozen, the absolute values of Softmax-GradCAMs are trapped within their initial small range (from $0$ to $10^{-3}$), and sigmoid probabilities stay near $0.5$. This prevents the convergence of cross entropy loss and weakens training robustness.

In this section, we attempted to scale CLIP logits or Softmax-GradCAMs to a reasonable range with a set of linear scaling parameters, which are adjusted manually or learned automatically. 
The results are shown in Table~\ref{tab:Loss Design App}, indicating that both manual and automatic scaling improved performance to some extent. However, our multi-label contrastive and CAM activation losses still achieved the best performance without introducing any additional parameters.
In addition, our CAM activation loss still outperforms pixel-wise cross entropy with scaled input by a large margin. This is because the CAM activation loss aligns the supervision signal and seed generation process.
During seed generation, Softmax-GradCAM is truncated with 0 and then subjected to min-max normalization. Then, values close to 1 are considered foreground candidates, while values close to 0 represent the background. 
Similarly, the CAM activation loss expects CAM values in the foreground to be close to the maximum response, i.e., close to 1 after normalization, while background values should be below 0.

\begin{table}[htpb]
\centering
\caption{
The performance evaluation of employing different multi-label classification and segmentation loss on PASCAL VOC 2012 \textit{trainaug} set. 
$\mathcal{L}_{BCE}$ stands for the binary cross entropy loss widely employed by other WSSS methods.
$\mathcal{L}_{CE}$ represents the pixel-wise cross entropy loss whose probabilities are obtained by inputting CAM activation values into a sigmoid function.
``Manual Scale" applies manually adjusted linear scaling parameters to scale the CLIP logits or Softmax-GradCAMs, and ``Auto Scale" scales values with learnable scaling parameters.
}
\label{tab:Loss Design App}
\scalebox{0.725}{
\begin{tabular}{cc|cc|cc|c}
\hline
$\mathcal{L}_{BCE}$ & $\mathcal{L}_{MCL}$ & $\mathcal{L}_{CE}$ & $\mathcal{L}_{CAL}$ & Manual Scale & Auto Scale &mIoU(\%) \\
\hline \hline
\checkmark & & & & & & 65.9 \\
\checkmark & & & & \checkmark &  & 70.7 \\
\checkmark & & & & & \checkmark & 70.1 \\
 & \checkmark & & & & & 71.5 \\
 & \checkmark & \checkmark & & & & 52.1 \\
 & \checkmark & \checkmark & & \checkmark & & 68.8 \\
 & \checkmark & \checkmark & & & \checkmark & 64.6 \\
 & \checkmark & & \checkmark & & & 74.2 \\
\hline
\end{tabular}
}
\end{table}

\section{Refine Full-Supervised Results with SAM-based Seeding Module}

\begin{table}[htpb]
\centering
\caption{
The effects of refining the \textbf{full-supervised} final segmentation results using SAMS or CRF on PASCAL VOC 2012 \textit{val} set.
}
\label{tab:SAMS Seg App}
\scalebox{0.9}{
\begin{tabular}{ll|l}
\hline
Backbone & Post-Processing & mIoU(\%) \\
\hline \hline
\multirow{3}{*}{DeepLab V3+} & - & 79.5 \\
 & CRF & 78.4{$_{\text{\color{ForestGreen}{-1.1}}}$} \\
 & SAMS & 80.9{$_{\text{\color{Maroon}{+1.4}}}$} \\
\hline
\multirow{3}{*}{Mask2Former} & - & 86.0 \\
 & CRF & 84.7{$_{\text{\color{ForestGreen}{-1.3}}}$} \\
 & SAMS & 85.5{$_{\text{\color{ForestGreen}{-0.5}}}$} \\
\hline
\end{tabular}
}
\end{table}

In Table~3 of the paper, attempts are made to refine the segmentation networks trained with seed by dense CRF or our SAM-based seeding module (SAMS). 
In this section, we conduct the same experiments on fully supervised conditions to demonstrate that the utility of SAMS is not limited to WSSS but can be applied to various semantic segmentation sub-tasks. 
The experimental results are presented in Table~\ref{tab:SAMS Seg App}.
On DeepLab V3+, SAMS achieves the same performance improvement as WSSS, while CRF remains ineffective. However, SAMS does not bring further improvements when employed to the high baseline results obtained by Mask2Former.

\section{More Implementation Details}
We implement our proposed method in PyTorch. All of our experiments are conducted on a single RTX 3090 GPU with 24GB memory. 
When training the proposed method for seed generation, we enable the second-order derivative to
ensure that the gradients of softmax-GradCAM are propagated correctly. Additionally, we detach the activation weights (Eq.~2 of paper) during softmax-GradCAM computation. 
For the final segmentation network, we train DeepLab V3+~\cite{Chen2018deeplab} with a batch size of 16 and Mask2Former~\cite{mask2former} of 4. 
Moreover, 120 epochs are trained on PASCAL VOC and 32 on MS COCO.
Other training and inference settings, such as the optimizer, scheduler, learning rate, etc., are set following implementations of MMSegmentation~\cite{mmseg2020}.

\begin{figure}[htpb]
\begin{center}
\includegraphics[scale = 0.14]{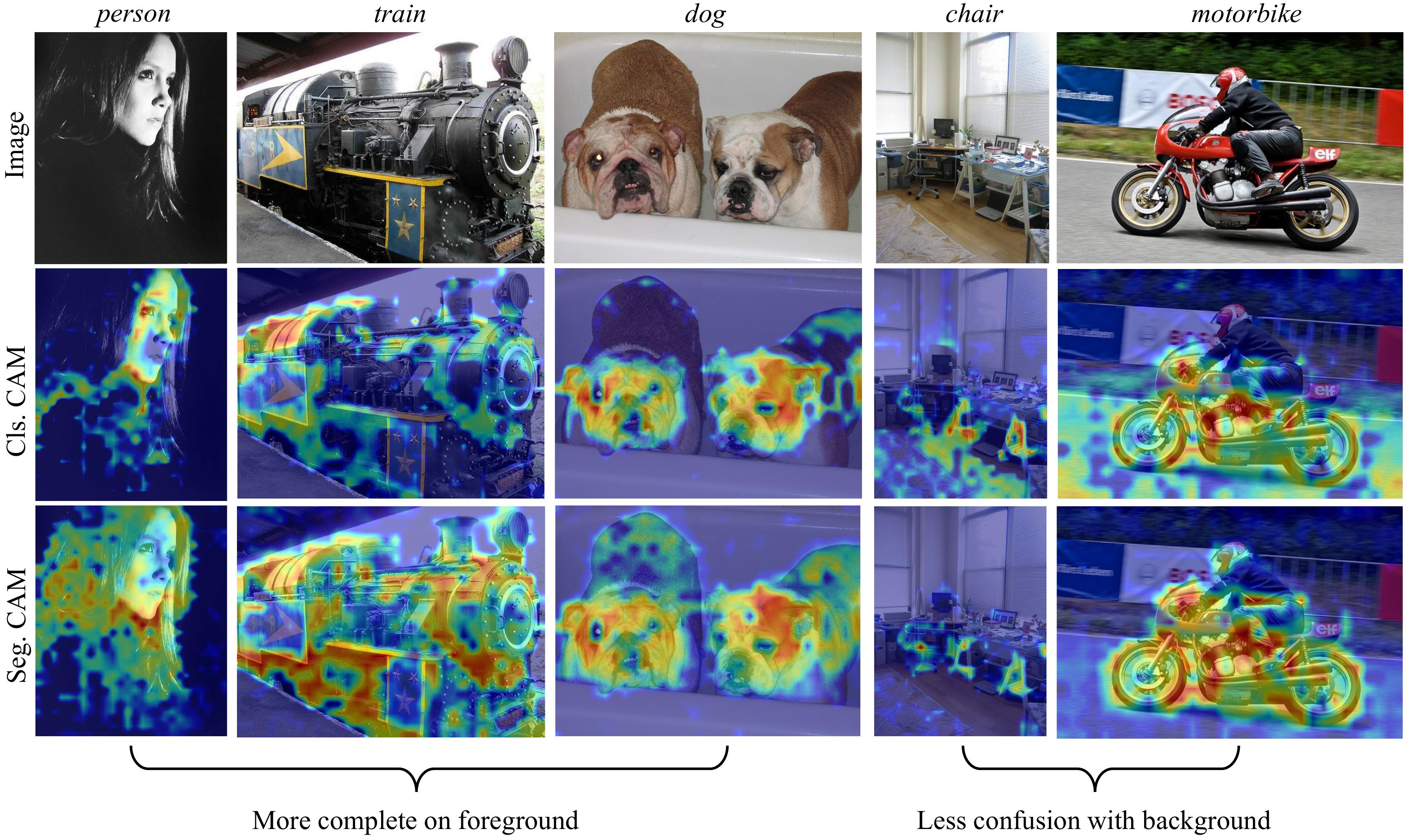}
\end{center}
\caption{
The GradCAMs generated by classification (Cls.) and segmentation (Seg.) tasks. The target class is
labeled on the top. Note that CAA are not employed for a clarity comparison.
}
\label{fig:compare_cams}
\end{figure}

\section{More Qualitative Results}
In Figure~\ref{fig:compare_cams}, we provide more results of GradCAMs generated by classification and segmentation tasks. We can observe that GradCAMs of the segmentation task are more complete and less likely to spread to background region, which proves the effectiveness of our coarse-to-fine design.

\begin{figure*}[htpb]
\begin{center}
\includegraphics[scale = 0.35]{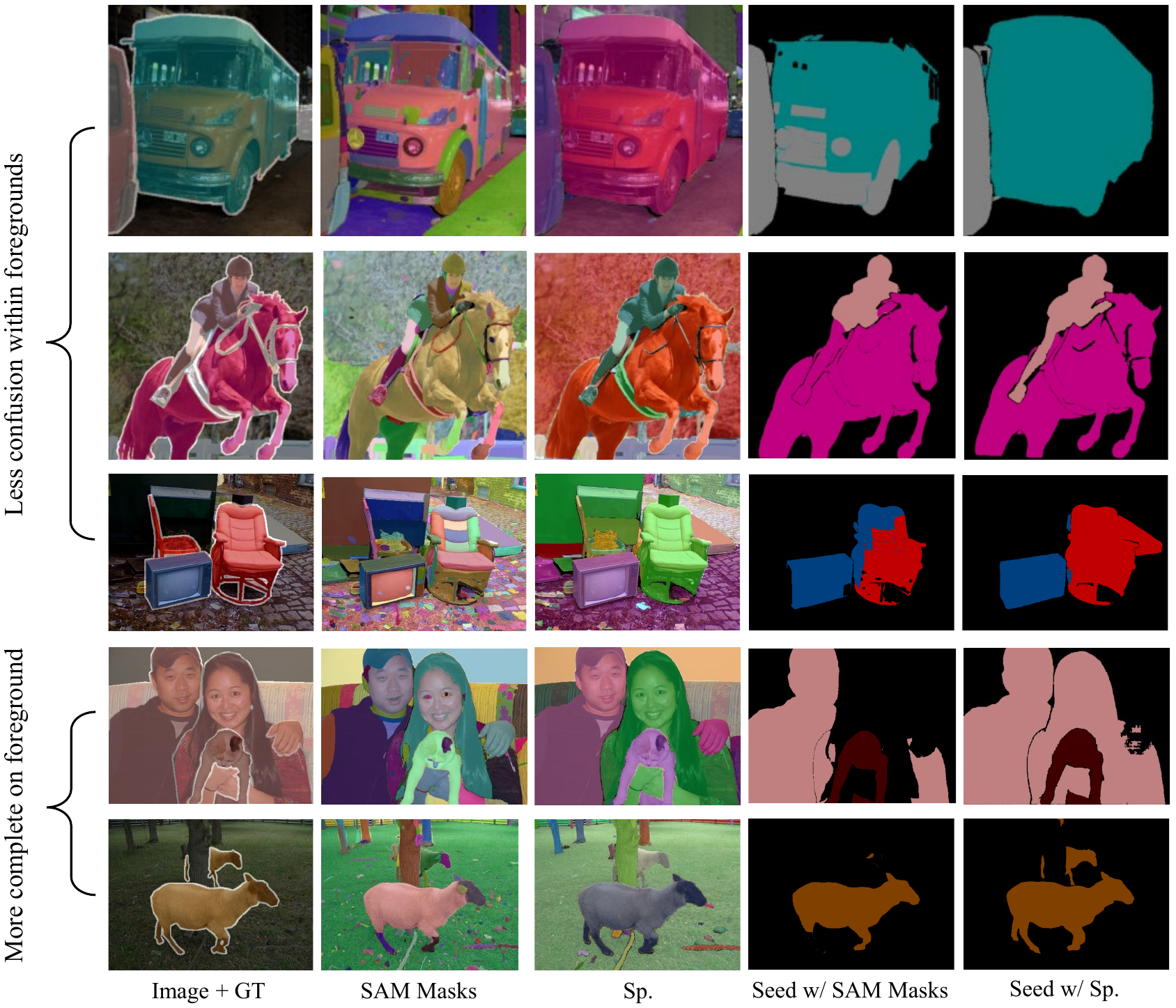}
\end{center}
\caption{
The seeds generated from original SAM masks and our SAM-based quasi-superpixel (Sp.).
}
\label{fig:compare_SAM_SAMS}
\end{figure*}

In Figure~\ref{fig:compare_SAM_SAMS}, we visualize the original SAM masks and our SAM-based quasi-superpixel, together with seeds generated base on them. 
It can be observed that our SAM-based quasi-superpixel tends to use as few masks as possible to identify the entire instance at once, with minimal overlap between masks. Therefore, seed based on quasi-superpixel is more effective at segmenting complete objects and ensuring consistent and accurate semantics within each instance.

\begin{figure*}[hpb]
\begin{center}
\includegraphics[scale = 0.3]{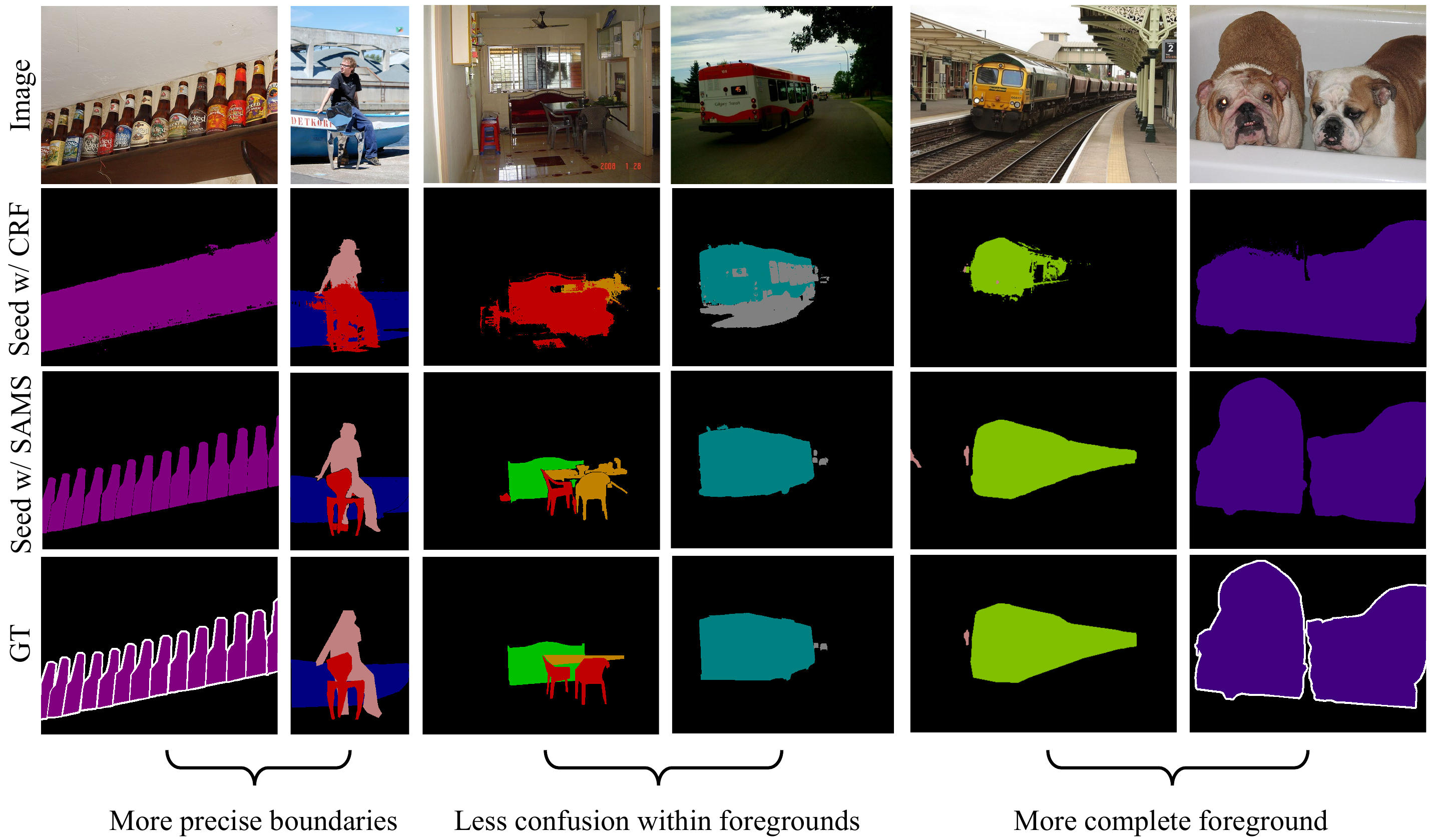}
\end{center}
\caption{
The seeds generated with different post-processing refinement methods, including dense CRF (CRF) and our SAM-based seeding module (SAMS).
}
\label{fig:compare_CRF_SAMS}
\end{figure*}

In Figure~\ref{fig:compare_CRF_SAMS}, we visualize the seeds generated from dense CRF and our SAM-based seeding module. 
We observe that SAMS can generate clearer and more accurate boundaries. 
When multiple foreground classes overlap and occlude each other, CRF is prone to confusion at the boundary, whereas SAMS avoids such confusion.
Finally, for elongated or color-variant objects, CRF often fails to propagate CAM throughout the entire object, while SAMS consistently identifies the complete object.

\end{document}